\documentclass{ecai} 

\usepackage{latexsym}
\usepackage{amssymb}
\usepackage{amsmath}
\usepackage{amsthm}
\usepackage{booktabs}
\usepackage{enumitem}
\usepackage{graphicx}
\usepackage{color}

\usepackage{todonotes}
\usepackage{cancel}
\usepackage{dsfont}

\newtheorem{theorem}{Theorem}
\newtheorem{lemma}[theorem]{Lemma}

\newtheorem{example}[theorem]{Example}

\newtheorem{definition}{Definition}

\newcommand{\BibTeX}{B\kern-.05em{\sc i\kern-.025em b}\kern-.08em\TeX}

\newcommand{\JL}[1]{
{#1}}

\newcommand{\FT}[1]{
{#1}}

\begin{document}


\begin{frontmatter}


\paperid{0879} 


\title{
Argumentation for Explainable Workforce 
Optimisation\\
(with Appendix)}


\author[A]{\fnms{Jennifer}~\snm{Leigh}}
\author[B]{\fnms{Dimitrios}~\snm{Letsios}}
\author[C]{\fnms{Alessandro}~\snm{Mella}} 
\author[C]{\fnms{Lucio}~\snm{Machetti}}
\author[A]{\fnms{Francesca}~\snm{Toni}\thanks{Corresponding Author. Email: f.toni@imperial.ac.uk.}}

\address[A]{Imperial College London, London, United Kingdom}
\address[B]{King's College London, London, United Kingdom}
\address[C]{Terranova, Florence, Italy}


\begin{abstract}
Workforce management is a complex problem involving the optimisation of the makespan and travel distance required for a team of operators to complete a set of jobs, using a set of instruments.
A crucial challenge in workforce management is accommodating changes at execution time so that explanations are provided to all stakeholders involved.
Here, we show that, by understanding workforce management
as abstract argumentation
in 
an industrial application,
we can accommodate change and obtain faithful  explanations.
We show, with a user study, that our tool and explanations lead to faster and more accurate problem solving than conventional manual approaches. 
\end{abstract}

\end{frontmatter}


\section{Introduction}

Workforce management is a complex problem amounting to optimising the makespan and distance required for a team of operators to travel and complete a set of jobs using a set of available instruments
.
It is a variant of the 
makespan scheduling problem~\cite{RefWorks:RefID:6-warner1972mathematical} combined with the travelling salesman problem (TSP) \cite{RefWorks:RefID:21-gutin2002traveling}, where
jobs have an associated processing time, location, skill 
and instrument requirements.
To complete these jobs, operators must have a start and end position, the set of skills required to complete the jobs, and any instruments
needed to complete the jobs.
Furthermore, instruments may require the operator to fulfill certain skill criteria. For example, suppose an operator is assigned a job to complete maintenance checks at 
some
site. The job may involve a car (the instrument), which would require the operator to have a driver's license (the skill)
. 

A crucial challenge in workforce management is accommodating changes at execution time so that explanations are provided to all stakeholders involved.
For example,
an operator may be assigned a job with an instrument that breaks down, or the operator may call in sick, or the job may be cancelled
.
Any such change may require an update to the overall schedule, with repercussions on other operators, jobs and instruments that need explaining to the stakeholders involved. For example, an operator may be given an extra job or a customer whose job is delayed may want to know why. However, it is not straightforward to provide an explanation that is both faithful and cognitively tractable, as in other optimisation settings~\cite{RefWorks:RefID:1-čyras2019argumentation}. 

Previous work \cite{RefWorks:RefID:1-čyras2019argumentation}  provide an abstract framework for explainability in the context of scheduling optimisation problems, based on
abstract argumentation (AA)~\cite{RefWorks:RefID:7-dung1995acceptability}
.
In this paper we develop a novel method for explainability in the context of workforce management optimisation problems, which is also based on AA.
As in 
\cite{RefWorks:RefID:1-čyras2019argumentation}, we use arguments to represent assignment decisions, where attacks indicate compatibility or incompatibility in the context of feasibility, efficiency and ability to satisfy additional constraints (\textit{fixed decisions}).
However, to accommodate the 
complexity of 
workforce management, our method addresses the following additional challenges:
(1) to accommodate sequential decisions in addition to assignment decisions, we use 
new attacks to check feasibility and efficiency of individual operator assignments; (2) to deal with efficiency beyond simply processing time, such as accounting for travel distance differences in our metric space, we use new AA frameworks which proxy travel impact using the triangle inequality, where attacks indicate inefficiencies and local search improvements for vehicle routing problems; (3) instruments are assigned to operators as well as jobs, leading to frameworks to satisfy feasibility; (4) we introduce constraints from skills and instruments, 
modeled using 
fixed decisions 
by \cite{RefWorks:RefID:1-čyras2019argumentation}.

Our novel method is designed to address industrial workforce management problems. 
Indeed, we have developed the method and a corresponding tool in collaboration with \emph{Terranova}\footnote{\url{https://www.terranovasoftware.eu/}}, a leading company providing advanced software solutions for the utilities sector, supporting digital innovation and process optimisation, and deployed it on \emph{Terranova}’s scheduling problems.
\JL{A key benefit offered by the tool to logistics operators is monitoring the progress of a
business process with jobs executed at different locations by operators and
involving vehicle movements. The tool supports
these manual modifications with explanations of whether a specific schedule update is
good or not and why.}
With the help of a user study with 28 \textit{Terranova} employees, we provide real evidence for applications of AA with optimisation in an industry setting, including evidence of efficiency and accuracy improvements. To our knowledge, this paper provides the first user study rationalising optimisation using argumentation. 

Overall, we make the following contributions:
(1) we provide a definition of the workforce management problem; (2) we map this problem onto AA \JL{with the aim to extract explanations}; (3) we define notions of explanations in this setting, \JL{relying upon the mapping onto AA}; and (4) we conduct a user study with expert industry players in this field, demonstrating the potential usefulness of our approach.

\FT{This pre-print is an extended version of 
\cite{paisversion}, including, in the Appendix,  proofs of some of the formal results.} 

\section{Related Work}

The workforce management problem 
shows similarities with the technician routing and scheduling problem (TRSP) \cite{gamst2024decision}, however the former introduces the notion of instrument requirement as well as skill, while the latter involves time windows for jobs and a penalty for any left uncompleted.
We adapt TRSP to define the additional parameter of instruments, where feasibility behaves similarly to job assignment, and fixed decision constraints similar to skills. The lack of time constraints and penalties leads to a 
generalised version of TRSP.

While our method's reliance on AA for explainability in optimisation is inspired by  \cite{RefWorks:RefID:1-čyras2019argumentation},
several other  works use argumentation for explainability in many settings \cite{vcyras2021argumentative}, 
including for ethics, medical informatics and security \cite{vassiliades2021argumentation}
%
 as well as for recommender systems \cite{RAGO2021103506}, though these rely upon different forms of argumentation.
 
\JL{Other instances of declarative AI have been used to solve scheduling problems like the workforce management problem. One example is Answer Set Programming (ASP), which offers alternative methodology to distribute workforce into teams with various skill constraints \cite{ricca2012team} and optimise travel routes \cite{gebser2018routing}. In particular, ASP proves useful for rescheduling when there are unexpected changes to the original problem at execution time \cite{cardellini2023rescheduling,dodaro2022operating}. While the above utilise symbolic AI/KR techniques to solve various scheduling problems, we only use AA to provide explanatory guidance to users at run time, and rely upon an underlying optimiser for initial solutions to the problem.
}

User studies are key to understanding the validity and usefulness of explanations. In particular, user studies provide insight into ``what makes an explanation \textit{meaningful} to a user" \cite{10.3389/frai.2024.1456486}. Various studies have been conducted to qualify the relationship between argumentative explanations and human-generated explanations, e.g. in \cite{naveed2018argumentation,sklar2018explanation,cerutti2021empirical}.
However studies such as these tend to be based on hypothetical scenarios rather than real-world applications and have rarely touched on optimisation problems.
Our user study is unique 
on 
assessing argumentative explanations for an industry problem.


\section{Background}

We give essential background on AA~\cite{RefWorks:RefID:7-dung1995acceptability} and its use explaining makespan scheduling~\cite{RefWorks:RefID:1-čyras2019argumentation}.

\begin{definition}\cite{RefWorks:RefID:7-dung1995acceptability}
\label{def: ArgumentationFramework}
    An \textbf{
    AA framework} $AF = (\mathrm{\textit{Args}}, \rightsquigarrow)$ is a directed graph where:
    \begin{itemize}
        \item \textit{Args} is a set of arguments (vertices), and
        \item $\rightsquigarrow$ is a binary ``attack" relation (directed edges) on \textit{Args} where $a \rightsquigarrow b$ means that $a$ attacks $b$.
    \end{itemize}
    For 
    $A,B \subseteq \mathrm{\textit{Args}}$, 
    $c \in \mathrm{\textit{Args}}$, let:
     $A \rightsquigarrow c $ iff $\exists d \in A$ such that $d \rightsquigarrow c$ and
     $A \rightsquigarrow B $ iff $\exists a \in A, b \in B : a \rightsquigarrow b$ (and $A \cancel{\rightsquigarrow} B$ means that it is not the case that $A \rightsquigarrow B$).
Then, an \textbf{extension} $E \subseteq \mathit{Args}$ is
    \begin{enumerate}
        \item \textbf{conflict-free} iff $E \cancel{\rightsquigarrow} E$,
        \item \textbf{stable} iff $E$ conflict-free and $E \rightsquigarrow a,  \forall a \in \mathit{Args} \backslash E$.
    \end{enumerate}
\end{definition}
\citet{RefWorks:RefID:1-čyras2019argumentation}
show how a variant of the makespan scheduling problem can be mapped to an AF as a route to explanation. In the  problem, we have $n$ independent jobs $\mathcal{J} = \{1, \ldots, n\}$ to be executed by $m$ operators $\mathcal{O} = \{1, \ldots, m\}$ and the aim is to minimise the total length of time any operator is working to complete their assigned jobs~\cite{RefWorks:RefID:5-graham1969bounds, RefWorks:RefID:6-warner1972mathematical}.
\emph{Solutions} to the makespan scheduling problem (i.e. optimal schedules), can be characterised by assignments of 1 or 0 to variables $x_{i,j}$, which represent the execution of job $j \in \mathcal{J}$ by operator $i \in \mathcal{O}$ \JL{with processing time $p_j$}.
We refer to any such solution as a matrix $(x_{i,j})$. \JL{For each operator, the sum of all processing times is the associated \textit{cost}, where $C_\mathrm{max} = \max_{i\in\mathcal{O}}\{C_i\}$ is the time for all operators to complete all jobs for a given solution.}

Solutions need to be \emph{feasible} (every job is assigned to one operator), and \emph{efficient} (the makespan cannot be reduced by moving jobs or swapping jobs between operators).
The makespan scheduling problem considered in \cite{RefWorks:RefID:1-čyras2019argumentation}
introduces 
 predetermined \emph{fixed decisions} to describe specific constraints a realistic problem may have:
 
\begin{definition}\cite{RefWorks:RefID:1-čyras2019argumentation}
\label{def: Fixed decisions D+ D-}
Let $D^-\subseteq \mathcal{J} \times \mathcal{O}$ and
$D^+\subseteq \mathcal{J} \times \mathcal{O}$
be, respectively, sets of  \textbf{negative
} and
\textbf{positive fixed decisions}.
Then, a solution $(x_{i,j})$
    satisfies fixed decisions, iff, for any  $i \in \mathcal{O}, j \in \mathcal{J}$: 
    
    $(i,j)\in D^- \implies x_{i,j} = 0, \quad
    (i,j)\in D^+ \implies x_{i,j} = 1.$
\end{definition}
Below, we recap these AFs
.
\begin{definition}
\label{def: updated feasibility AF}
    \cite{RefWorks:RefID:1-čyras2019argumentation}
    The \textbf{feasibility AF} is $AF_F \!=\! (\mathit{Args_F}, \rightsquigarrow_F)$ with
    \begin{itemize}
        \item $\mathit{Args_F} = \{a_{i,j} : i\in \mathcal{O}, j\in \mathcal{J}\}$,
        \item $a_{i,j} \rightsquigarrow_F a_{k,l} \iff i\neq k, j = l$.
    \end{itemize}
    \label{def: optimality AF}
The \textbf{optimality AF} is $(\mathit{Args_S}, \rightsquigarrow_S)$ with $\mathit{Args_S} = \mathit{Args_F}$ and:
    \begin{itemize}
        \item $\rightsquigarrow_S = (\rightsquigarrow_F \backslash \{(a_{i,j}, a_{i',j'}):C_i = C_\mathrm{max}, x_{i,j} = 1, C_i > C_{i'} + p_j\}) \cup \{(a_{i',j'}, a_{i,j}):C_i = C_\mathrm{max}, x_{i,j} = 1, x_{i',j'} = 1, i \neq i', j \neq j', p_j > p_{j'}, C_i + p_{j'}> C_{i'} + p_j\}$.
    \end{itemize}
    \label{def: Fixed Decision AF}
 For 
 fixed decisions $D \!=\! 
 (D^-, D^+)$, the \textbf{fixed decision AF} is $(\mathit{Args_D}, \rightsquigarrow_D)$ with $\mathit{Args_D} = \mathit{Args_F}$ and:
    \begin{itemize}
        \item $\rightsquigarrow_D = (\rightsquigarrow_F \cup \{(a_{i,j}, a_{i,j}): (i,j) \in D^-\}) \backslash \\ \{(a_{k,l}, a_{i,j}): (i, j) \in D^+, (k,l) \in \mathcal{O} \times \mathcal{J} \}$.
    \end{itemize}
\end{definition}

The feasibility AF ensures jobs are assigned exactly once
. The optimality AF identifies violations of the single exchange property (where moving one job to another operator will reduce the total maximum processing time of any operator) and pairwise exchange property (where swapping two jobs between two operators will reduce the total maximum processing time). The fixed decision AF enforces any additional constraints – self-attacks signify a violation of negative fixed decisions, while non-attacks are due to positive fixed decisions.

\citet{RefWorks:RefID:1-čyras2019argumentation} prove that all three AFs can be constructed in $O(nm^2)$ time and verification of whether an extension is stable in each of these AFs requires up to $O(n^2m^2)$ time. Furthermore, 
stable extensions in each of these  AFs correspond to schedules with the intended properties in the makespan scheduling problem, and can be used as a basis for explanation.
{For a schedule $S$, we use $E\approx S$ to refer to the corresponding extension.  }

\section{Workforce Management Problem}
\label{sec:prob}

A problem instance is a triple $\langle \mathcal{O},\mathcal{J},\mathcal{I}\rangle$,
where $\mathcal{O}$ is a set of operators, $\mathcal{J}$ is a set of jobs, and  $\mathcal{I}$ is a set of instruments.
Each job $j\in\mathcal{J}$ corresponds to a task (e.g.\ equipment maintenance or process control) that needs to be implemented at a specific location (e.g.\ manufacturing site or customer location) by exactly one operator.
The goal is to decide (1) an assignment of jobs to operators and (2) a sequence of the jobs assigned to each operator subject to requirements:

\begin{itemize}
\item \textbf{Makespan Objective.}
Operator $i\!\in\!\mathcal{O}$ needs $p_{i,j}$ units of time to complete job $j\!\in\!\mathcal{J}$.
Also, $d_{j,j'}$ is the distance between the locations of 
jobs  $j,j'\!\in\!\mathcal{J}$.
The \emph{makespan}, i.e.\ the total processing and traveling distance of each operator, should be minimised.

\item \textbf{Restricted Assignments.}
Implementing job $j\in \mathcal{J}$ requires using a subset $\mathcal{I}_j^{\mathcal{J}}\subseteq\mathcal{I}$ of instruments. 
Operator $i\in \mathcal{O}$ is qualified to use a subset $\mathcal{I}_i^{\mathcal{O}}$ of the available instruments. Job $j\in\mathcal{J}$ can be assigned to operator $i\in \mathcal{O}$ only if $\mathcal{I}_i^{\mathcal{O}}\subseteq\mathcal{I}_j^{\mathcal{J}}$.
\end{itemize}

The problem is defined in terms of three 
sets: \emph{jobs} $\mathcal{J}$ (with $n = |\mathcal{J}|$), \emph{operators} $\mathcal{O}$  (with $m = |\mathcal{O}|$) and \emph{instruments} $\mathcal{I}$  (with $t = |\mathcal{I}|$).
These sets are characterised by the following quantities:
\begin{definition}
    \label{def: x_ij z_itau}
    Define $x_{i,j,k} = 1$ iff job $j \in \mathcal{J}$ is the $k$th job allocated to operator $i \in \mathcal{O}$, and $0$ otherwise.
    Similarly, $z_{i, \tau} = 1$ iff instrument $\tau \in \mathcal{I}$ is allocated to operator $i \in \mathcal{O}$, and $0$ otherwise.

    \JL{Let $z_j$ indicate if a job is processed last by some operator. Then $z_j = 1$ iff $\exists i \in \mathcal{O}, j \in \mathcal{J}, k\in [n-1]: x_{i,j,k} = 1, \sum_{j’} x_{i,j’,k+1} = 0$, and $0$ otherwise.}
\end{definition}

\begin{definition}
\label{def: processing times p_ij}
 $P \in M_{m, n}(\mathds{R})$ is the $m \times n$ \textbf{processing times matrix} for  
 $\mathcal{J}$ and $\mathcal{O}$. For  
 $i \in \mathcal{O}$ and $j \in \mathcal{J}$,
 $p_{i,j}$ is the time taken for operator $i$ to complete job $j$.
\end{definition}

\begin{definition}
    \label{def: new distance metric}
    Let $d_{j,j'}$ be the distance between $j, j'\in \mathcal{J}$.
    \JL{$d_{0, j}=d_{j,0}$ is the distance between the origin $(0,0,0)$ and job $j$.}
\end{definition}
We take $d$ to be Euclidean distance and the start/end point as $(0,0)$.

\begin{definition}
\label{def: binary indicator for job order scheduling}
Let $i\in\mathcal{O}$ be an operator and $j,j'\in\mathcal{J}$ be two jobs. Define $y_{i,j,j'}=1$ iff both $j$ and $j'$ are assigned to $i$, and $j'$ should be completed immediately after $j$ (with $y_{i,j,j'}=0$ otherwise).
\end{definition}

\begin{definition}
\label{def: ExtendedCostSchedulingProblem}
The \textbf{workforce management optimisation problem} is:
\JL{
\begin{equation*}
\min_{C_\mathrm{max}, C_i, x_{i,j,k}, y_{i,j,j'}} C_\mathrm{max}
\end{equation*}
subject to}
\begin{equation*}
\begin{array}{rl}
    C_\mathrm{max} \geq C_i, & i\in \mathcal{O} \\
    C_i = \alpha \sum_{j,k} x_{i,j,k} p_{i,j} + \beta \big(\sum_{j,j'} y_{i,j,j'} d_{j,j'}
    & \\
    + \JL{\sum_{j} x_{i,j,0} d_{0,j} + \sum_{j} z_{j} d_{j,0}} \big),
    & i \in \mathcal{O} \\ 
    y_{i,j,j'} \geq \JL{x_{i,j,k} + x_{i,j,k+1}},  & i \in \mathcal{O},  \\&j,j' \in \mathcal{J}, \\&k \in [n-1] \\
    \sum_{i,k} x_{i,j,k} = 1,  & j \in \mathcal{J} \\
    \sum_{j} x_{i,j,k+1} \leq \sum_{j} x_{i,j,k},  & i \in \mathcal{O}, \\&k \in [n-1] \\
    \sum_{j} x_{i,j,k} \leq 1,  & i \in \mathcal{O}, k \in [n]\\
    \JL{z_j \geq x_{i,j,k} - \sum_{j’} x_{i,j’,k+1}}& \JL{i \in \mathcal{O},} \\&\JL{k \in [n-1]}
    
\end{array}
\end{equation*}
where $C_\mathrm{max} = \max_{i\in\mathcal{O}} \{C_i\}$
and $\alpha, \beta \in \mathbb{R}$ take arbitrary values
such that $\alpha + \beta = 1$. (In this study, we use $\alpha = \beta = 0.5$.)
\end{definition}

The first two equations describe the objective function: to minimise the maximum cost of each operator. Cost, as defined in the third equation, is the sum of all processing times for each operator and the sum of distances for each operator to travel between jobs. $\alpha$ and $\beta$ reflect the weighting of these cost quantities.
The forth equation indicates the distance between two jobs should only be included in the cost function if these jobs are sequential and assigned to the same operator.
The notation $y_{i,j,j'}$ follows from problems such as TSP \cite{gouveia2017extended, RefWorks:RefID:21-gutin2002traveling}.
The fifth equation indicates that every job must be assigned to exactly one operator in exactly one position.
The sixth equation is a position-dependent formulation constraint, ensuring an operator's jobs must be sequential. 
The seventh equation indicates every operator's schedule can have at most one job in each position. The last constraint accounts for operators which do not have the same number of jobs, where $z_j$ indicates the last job assigned to an operator.

\begin{example}
    Consider operator $i$ starts and ends at 
    (0,0). Let the coordinates of job $j$ be $(3, 4)$ and $p_{i,j} \!=\! 3$. Then $x_{i,j,1} p_{i,j} \!=\! 3$.
    Suppose $x_{i,j,1}\!=\!1$ and $x_{i,j',k}\!=\!0, \forall j'\neq j, k\neq 1$.
    Then
     $\sum y_{i,j,j'}d_{j,j'} = 0$.
     Lastly, we have $d_{i, 0} = d_{i,N} = \sqrt{3^2+4^2} = 5$. Putting this together with $\alpha=\beta=0.5$, $C_i = 0.5(3) + 0.5 (5 + 5) = 6.5.$
\end{example}

\begin{example}
\label{ex:efficiency AF example}
    Consider the workforce management optimisation problem consisting of three jobs, with locations $(3, 4), (5, 12)$ and $ (5, 12)$ respectively, and two operators, both beginning and ending their shifts at the origin $(0, 0)$.
    Let $p_{i,j}$ and schedule $S$ be:
    $$p_{i,j} = \begin{pmatrix}
        120 & 60 & 30\\
        120 & 60 & 60
    \end{pmatrix}_{i,j} \quad
   S = \begin{pmatrix}
        1 & 0 & 1\\0& 1& 0
    \end{pmatrix}.$$
    To calculate the maximum cost of the problem, $C_{max}$, we must calculate the cost of each operator.
    \JL{We have $C_1 = 0.5 (120+30)+0.5 (5+\sqrt{68}+13) \approx 88.12, C_2 = 0.5 (60) + 0.5(13 + 13) = 43.$
    So $C_{max} = C_1 \approx 88.12$.}
    This means operator 1 has the highest cost and thus this is the total cost of the solution. Any changes to reduce the total cost must be applied to operator 1.
    \end{example}
We will refer to a solution of the workforce management optimisation problem as an \emph{optimal schedule wrt extended cost}.
\section{Argumentative Workforce Management}
\label{sec:theory}
Inspired by \cite{RefWorks:RefID:1-čyras2019argumentation}, we map our problem onto AFs.
However, in contrast to \cite{RefWorks:RefID:1-čyras2019argumentation}, we accommodate travel distance, impacting how we define efficient solutions (Subsection \ref{subset: extended cost & efficiency}). We consider if the order of jobs assigned to each operator can be optimised to minimise the total distance travelled (Subsection \ref{subsec: individual efficiency}). Finally, we consider the impact of skills and instruments on fixed decisions (i.e. what operator-job-instrument combinations are impossible)  (Subsections \ref{subsec: Mapping Skill Constraints to Fixed Decisions} and \ref{subsec: instruments}).

To accommodate job sequencing, we use AA to check efficiency, compared to both other operators and variations of the same operator's schedule.
We do not use AA methods for sequence feasibility checks (i.e. all jobs assigned to exactly one place in the sequence, and each place in a sequence has an assigned job).

\subsection{Extended Cost \& Efficiency}
\label{subset: extended cost & efficiency}
We first define suitable notions of exchange properties to deal with fixed decisions, when focusing on efficiency of solutions.  
\begin{definition}
\label{def: SEP+ PEP+ extended cost efficient}
    Consider a schedule $S$ and \emph{critical operator} $i\in\mathcal{O}$, i.e.\ $C_i = C_\mathrm{max}$.
    Given $j \in \mathcal{J}$ executed by $i$ at position $k$, we define: 
    \begin{enumerate}
        \item \textbf{Extended Single Exchange Property} (SEP+):
        
        \JL{$\forall i'\in\mathcal{O}, j', j'^+ \in \mathcal{J},x_{i',j',k'}=x_{i',j'^+,k'+1}=1$:}
        \JL{$C_i - C_{i'} \leq \alpha p_{i',j} + \beta [d_{j,j'} + d_{j, j'^+}] - \beta d_{j',j'^+} $}
        
        \item \textbf{Extended Pairwise Exchange Property} (PEP+):
        
        $\forall j^{-}, j^{+}, j^{'-},j^{'+}\in \mathcal{J}:x_{i,j^{-},k-1}=x_{i,j^{+},k+1}=x_{i',j^{'-},k'-1}=x_{i',j^{'+},k'+1}=1 :$ 
        $\beta ([d_{j^-, j} + d_{j, j^+}] - [d_{j^-, j'} + d_{j', j^+}]) > \alpha [p_{i, j'}-p_{i,j}] $
         $\implies$ 
         $ C_i - C_{i'} \leq
\alpha [p_{i', j}-p_{i',j'}] + \beta (d_{j'^-, j} + d_{j, j'^+}] - [d_{j'^-, j'} + d_{j', j'^+}]).$
    \end{enumerate}
    $S$ is \textbf{extended cost efficient} if $S$ is feasible, satisfies SEP+ and PEP+.
\end{definition}
    Intuitively, 
    $S$ satisfies SEP+ if the cost $C_{max}$ cannot be decreased by moving a job $j$ from critical operator $i$ to non-critical operator $i'$. Also, $S$ satisfies PEP+ if $C_{max}$ cannot be decreased by swapping the assigned operator and position of 
    $j, j'$ where $x_{i,j,k}=x_{i',j',k'}=1$. \JL{Note, pairwise exchange is defined as the swapping of 2 jobs, as opposed to a crossover of schedules.}
    \JL{Full explanations can be found in the Appendix.}
    
\begin{lemma}
\label{lemma: Every optimal schedule with respect to extended cost satisfies SEP+ and PEP+}
    Every optimal schedule wrt extended cost satisfies SEP+ and PEP+.
\end{lemma}
The proof of this result can be found in the Appendix. 
We then adapt the definition of optimality AF in
~\cite{RefWorks:RefID:1-čyras2019argumentation}, applying attacks where arguments violate the SEP+ and PEP+ conditions
.

\begin{definition}
    \label{def: extended cost efficiency AF}
    For feasibility AF $(\mathit{Args_F}, \rightsquigarrow_F)$ and schedule $S$, the \textbf{extended cost efficiency AF} $(\mathit{Args_{S+}}, \rightsquigarrow_{S+})$ is defined as:
    \begin{itemize}
        \item $\mathit{Args_{S+}} = \mathit{Args_F}$,
        \item 
        $\rightsquigarrow_{S+} = \bigl(\rightsquigarrow_F \backslash \bigl\{(a_{i,j}, a_{i',j}):C_i = C_\mathrm{max}, x_{i,j,k} = 1: \exists j' \neq j, x_{i',j'}=1 :$
        $C_i - C_{i'} > p_{i',j} + \beta [d_{j', j} + d_{j, j'^+}] - \beta d_{j', j'^+}\bigr\}\bigr) \bigcup \\ \bigl\{(a_{i',j'}, a_{i,j}):C_i = C_\mathrm{max}, x_{i,j,k} = x_{i',j', k'} = 1, i \neq i', j \neq j', 
        \beta \bigl([d_{j^-, j} + d_{j, j^+}] - [d_{j^-, j'} + d_{j', j^+}]\bigr) > p_{i, j'}-p_{i,j}, \\
         C_i - C_{i'} > p_{i', j}-p_{i',j'} + \beta \bigl([d_{j'^-, j} + d_{j, j'^+}] - [d_{j'^-, j'} + d_{j', j'^+}]\bigr)
        \bigr\}.$
    \end{itemize}
    \end{definition}
    In other words: (i) if an argument $a_{i,j} \in E$ violates SEP+ due to $a_{i',j}$, remove any attacks from $a_{i,j}$ to $a_{i',j}$;
        (ii) if arguments $a_{i,j}, a_{i',j'} \in E$ cause a violation of PEP+, add an attack from $a_{i',j'}$ to $a_{i,j}$.
Note the criteria for self-attacks is SEP+ and the criteria for attacks between two different arguments is PEP+, both as defined in Definition \ref{def: SEP+ PEP+ extended cost efficient}.
With an abuse of notation, we will refer to the extended cost efficiency AF simply as $(\mathit{Args_{S}}, \rightsquigarrow_{S})$ for ease.

\begin{theorem}
\label{theo:feas}
    {For a 
    schedule $S$ and $S \approx E$, 
    $E$ is stable in $(\mathit{Args_{S}}, \rightsquigarrow_{S})$ iff $S$ is feasible and satisfies SEP+ and PEP+.}  
\end{theorem}
The proof of this result can be found in the Appendix. 

\begin{example}
\label{ex: efficiency AF example with SEP+/PEP+ calculations}
    Consider the workforce management optimisation problem in Example~\ref{ex:efficiency AF example}.
    We can do the following checks to build our extended cost efficiency AF:
    \begin{itemize}
        \item Move job 1 from operator 1 to 2 at start:
        $C_1 - C_2 = 44.85-35.6 = 9.25$,
        $p_{i',j} + [d_{j', j} + d_{j, j'^+}] - \beta d_{j', j'^+} = 12.9 + [5+\sqrt{68}]-13 \approx 13.15$.
Then, 
        $9.25 < 13.15 \implies$ satisfies SEP+.
        \item Move job 1 from operator 1 to 2 after job 2: same as above.
        \item Move job 3 from operator 1 to 2 at start:        
        $p_{i',j} + [d_{j', j} + d_{j, j'^+}] - \beta d_{j', j'^+} = 8.7 + [13 + 0]-13 = 8.7$.
Then, 
        $C_1 - C_2 = 9.25 > 8.7 \implies$ violates SEP+.

        \item Swap jobs 1 and 2:
        $p_{i, j'}-p_{i,j} = 0$,
        $\beta ([d_{j^-, j} + d_{j, j^+}] - [d_{j^-, j'} + d_{j', j^+}]) = [5+\sqrt{68}] - [13+0] \approx 0.246 > 0$,
        $C_i - C_{i'} = 9.25$,
        $p_{i', j}-p_{i',j'} + \beta \bigl([d_{j'^-, j} + d_{j, j'^+}] - [d_{j'^-, j'} + d_{j', j'^+}] = 12.9 - 9.6 + ([5 + 5] - [13 + 13] = -12.7 < 9.25 \implies$ violates PEP+.

        \item Swap jobs 3 and 2:
        $p_{i, j'}-p_{i,j} = 9.6 - 5.7 = 3.9$,
        $\beta ([d_{j^-, j} + d_{j, j^+}] - [d_{j^-, j'} + d_{j', j^+}]) = [\sqrt{68} + 13] - [\sqrt{68} + 13] = 0 < 3.9 \implies$ satisfies PEP+.
    \end{itemize}
    The corresponding efficiency AF is shown in Figure \ref{fig:efficiency AF example with SEP+/PEP+ calculations}.
    
    \begin{figure}
        \centering
        \includegraphics[width=0.35\linewidth]{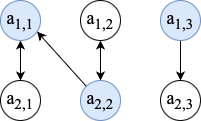}
        \caption{ Efficiency AF for Example \ref{ex: efficiency AF example with SEP+/PEP+ calculations}}
        \label{fig:efficiency AF example with SEP+/PEP+ calculations}
    \end{figure}
\end{example}
\begin{lemma}
\label{lemma: time complexity extended cost efficiency AF}
    {The extended cost efficiency AF $(\mathit{Args_{S}}, \rightsquigarrow_{S})$ can be constructed in $O(m^2n^2)$ time.
    Given a schedule $S$,
    verifying whether 
    extension $E \approx S$ is stable can be done in $O(m^2n^2)$ time.}
    
\end{lemma}
The proof of this result can be found in the Appendix. 

\subsection{Individual Efficiency}
\label{subsec: individual efficiency}
When considering distance in the workforce management problem, the order of jobs should be optimised. Consider three jobs, where jobs 1 and 2 occur at the same location and job 3 at a separate location. Completing the first two jobs, then the third, would be more efficient than completing job 1, then job 3, then job 2.

\begin{definition}
\label{def: ISEP IPEP individual cost efficient}
    Let $d$ be the distance metric.
    For $ x_{i,j,k} = x_{i,j',k'} = 1$, we define the following properties $\forall j\neq j', k\neq k'$:
    \begin{enumerate}
        \item \textbf{Individual Single Exchange Property} (ISEP): $\forall x_{i,j^-,k-1}=x_{i,j^+,k+1}=x_{i,j'^+,k'+1}=1 :$
        
        $d_{j', j} + d_{j, j'^+} - d_{j', j'^+} \geq d_{j^-, j} + d_{j, j^+} - d_{j^-, j^+}$
        \item \textbf{Individual Pairwise Exchange Property} (IPEP): 
        $\forall x_{i,j^-,k-1}=x_{i,j^+,k+1}=x_{i,j'^-,k'-1}=x_{i,j'^+,k'+1}=1 :
            d_{j'^-, j} + d_{j, j'^+} + d_{j^-, j'} + d_{j', j^+} \geq d_{j'^-, j'} + d_{j', j'^+} + d_{j^-, j} + d_{j, j^+}.$
    \end{enumerate}

    We say $S$ is \textbf{individual cost efficient} iff $S$ is feasible and satisfies both ISEP and IPEP.
\end{definition}

\begin{lemma}
\label{lemma:optimal schedule satisfies ISEP IPEP}
    Every optimal schedule with respect to individual cost satisfies ISEP and IPEP.
\end{lemma}
\JL{The proof follows the same approach as Lemma \ref{lemma: Every optimal schedule with respect to extended cost satisfies SEP+ and PEP+}}.

\begin{example}
\label{ex: individual efficiency}
    Consider an operator $i$ with 3 jobs at locations 1$(3, 4)$, 2$(5, 12)$ and 3$(5, 12)$,
    and the operator beginning and ending its shift at $(0, 0)$. Suppose the individual schedule order is $\{2, 1, 3\}$. We can then check the following (not a conclusive list):
    \begin{itemize}
        \item Move job 1 to start:
        $d_{j', j} + d_{j, j'^+} - d_{j', j'^+} = 5 + \sqrt{68} - 13 \approx 0.246$,
        $d_{j^-, j} + d_{j, j^+} - d_{j^-, j^+} = 13 + 13 - 0 = 26 > 0.246 \implies$ violates ISEP.
        
        \item Swap jobs 2 and 3 ($j' = 2, j = 3$):
        $d_{j'^-, j} + d_{j, j'^+} + d_{j^-, j'} + d_{j', j^+} = 13 + \sqrt{68} + \sqrt{68} + 13 \approx 42.49$,
        
        $d_{j'^-, j'} + d_{j', j'^+} + d_{j^-, j} + d_{j, j^+} = 13 + \sqrt{68} + \sqrt{68} + 13 \approx 42.49 \implies$ satisfies IPEP.

        \item Swap jobs 1 and 3 ($j' = 1, j = 3$):
        $d_{j'^-, j} + d_{j, j'^+} + d_{j^-, j'} + d_{j', j^+} = 0 + 0 + 0 + 5 = 5$,
        
        $d_{j'^-, j'} + d_{j', j'^+} + d_{j^-, j} + d_{j, j^+} = \sqrt{68} + \sqrt{68} + \sqrt{68} + 13 \approx 37.74 > 5 \implies$ violates IPEP.
    \end{itemize}
\end{example}

\begin{definition}
    \label{def: individual efficiency AF}
    For feasibility AF $(\mathit{Args_F}, \rightsquigarrow_F)$ and schedule $S$, the \textbf{individual cost efficiency AF} $(\mathit{Args_{IS}}, \rightsquigarrow_{IS})$ is defined as:
    \begin{itemize}
        \item $\mathit{Args_{IS}} = \mathit{Args_F}$,
        \item 
        $\rightsquigarrow_{IS} = \rightsquigarrow_F \bigcup \bigl\{(a_{i,j}, a_{i,j}):\exists w \neq j, x_{i,w}=1 : \\ d_{j', j} + d_{j, j'^+} - d_{j', j'^+} < d_{j^-, j} + d_{j, j^+} - d_{j^-, j^+}\bigr\} \\ \bigcup \bigl\{(a_{i,j'}, a_{i,j}): x_{i,j} = x_{i,j'} = 1, j \neq j',
            d_{j'^-, j} + d_{j, j'^+} + d_{j^-, j'} + d_{j', j^+} < d_{j'^-, j'} + d_{j', j'^+} + d_{j^-, j} + d_{j, j^+}
        \bigr\}.$
    \end{itemize}
\end{definition}

\begin{example}
    \label{ex: individual efficiency af}
    Figure \ref{fig: individual efficiency AF} gives the final AF for 
    Example \ref{ex: individual efficiency}.
    Note 
    that $a_{1,1}\rightsquigarrow a_{1,1}$ due to ISEP violations and 
    $a_{1,1}\rightsquigarrow a_{1,2}$,
    $a_{1,2}\rightsquigarrow a_{1,1}$,
    $a_{1,1}\rightsquigarrow a_{1,3}$, $a_{1,3}\rightsquigarrow a_{1,1}$ due to IPEP violations.
    
    \begin{figure}[h]
        \centering
        \includegraphics[width=0.33\linewidth]{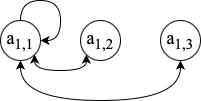}
        \caption{Individual cost efficiency AF for Example \ref{ex: individual efficiency af} }
        \label{fig: individual efficiency AF}
    \end{figure}
\end{example}

\begin{lemma}
\label{lemma: time complexity individual cost efficiency AF}
    {
    The individual cost efficiency AF $(\mathit{Args_{IS}}, \rightsquigarrow_{IS})$ can be constructed in $O(nm(n+m))$ time.
    Given a schedule $S$, verifying whether 
    extension $E \approx S$ is stable can be done in $O(m^2n^2)$ time.
    }
\end{lemma}

\JL{The proof follows the same approach as Lemma \ref{lemma: time complexity extended cost efficiency AF}}.

\subsection{Skill Constraints to Fixed Decisions}
\label{subsec: Mapping Skill Constraints to Fixed Decisions}

In workforce management, often not every operator is equipped to complete every job. Some jobs may 
require more skills than others – for example, an engineer may have a different set of skills to a van driver. In the next two subsections, we consider three types of fixed decisions that should be fulfilled: the first is meeting the skills requirements to complete a given job.

\begin{definition}
\label{def: skill prerequisites}
    Let $\mathcal{K}$ be a set of \textbf{skills}, $\mathcal{K} = \{k_1, k_2, \ldots, k_K \}$ where $|\mathcal{K}| = K$. Then the set of \textbf{skill prerequisites} is $\mathcal{K}_j \subseteq \mathcal{K}$ for a job $j \in \mathcal{J}$ and the set of \textbf{operator skills} is $\mathcal{K}_i \subseteq \mathcal{K}$ for an operator $i \in \mathcal{O}$.
    By definition, any operator $i$ assigned a job $j$ must fulfill all skill criteria. In other words, a feasible (with relation to fixed decisions) schedule $S$ must satisfy $\mathcal{K}_j \subseteq \mathcal{K}_i
    $.

    $\mathcal{I}_j \subseteq \mathcal{I}$ is the set of instrument requirements for job $j \in \mathcal{J}$.
\end{definition}
\begin{definition}[Based on {\v{C}}yras et al.'s~\cite{RefWorks:RefID:1-čyras2019argumentation} Fixed Decision AF]
\label{def: skill constraints AF}
    The \textbf{skill constraints AF} $(\mathit{Args_{D_S}}, \rightsquigarrow_{D_S})$ is defined as: 
    \begin{itemize}
        \item $\mathit{Args_{D_S}} = \mathit{Args_F}$,
        \item $\rightsquigarrow_{D_S} = (\rightsquigarrow_F \cup \{(a_{i,j}, a_{i,j}): \exists k \in \mathcal{K}_j, k \notin \mathcal{K}_i\})$.
    \end{itemize}
\end{definition}
\JL{Using Definitions \ref{def: Fixed decisions D+ D-}, \ref{def: skill prerequisites} and \ref{def: skill constraints AF}, it follows that:}
\begin{lemma}
\label{lemma: violating skill requirements are negative fixed decisions}
Violating skill requirements are negative fixed decisions. In other words:
    $\exists k \in \mathcal{K}_j : k \notin \mathcal{K}_i \implies (i, j) \in D^-.$
\end{lemma}
\begin{example}
\label{ex: skill requirement af}
    Consider a problem consisting of 3 operators and 3 jobs. Let operator 1 have skills (A, B, C), operator 2 have skills (A, C) and operator 3 have skills (B, C). Similarly, let the skill requirements for jobs be job 1 (A), job 2 (B) and job 3 (B, C).
    \begin{figure}
            \centering
            \includegraphics[width=0.4\linewidth]{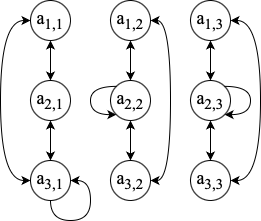}
            \caption{ Skill requirement AF for Example \ref{ex: skill requirement af}}
            \label{fig: skill requirement af}

        \end{figure}
    So $D^- = \{(2, 2), (2, 3), (3, 1)\}$. Figure \ref{fig: skill requirement af} shows the corresponding skill requirements (fixed decision) AF.
\end{example}
\begin{lemma}
\label{lemma: time complexity skill constraint AF}
   {Given 
   a set of job skill prerequisites and a set of operator skills, the skill constraints AF can be constructed in $O(mnK + nm^2)$ time.
    Given a schedule $S$, verifying whether an extension 
    $E \approx S$ is stable for $AF_S$ can be done in $O(n^2m^2)$ time.}
\end{lemma}
The proof is similar to
those of earlier lemmas
.

\subsection{Instrument Allocations}
\label{subsec: instruments}
Some jobs may require additional instruments to complete a job, like a toolkit or a van. Instruments may even require additional skills, such as a driver's license
. Like jobs, instruments must be allocated to exactly one operator, assuming the operator 
has the correct skills.
\begin{definition}
    \label{def: instruments}
    $\mathcal{I}$ is the set of instruments.
    $\mathcal{K}_\tau \subseteq \mathcal{K}$ is the set of skill requirements for instrument $\tau \in \mathcal{I}$.
\end{definition}
\begin{lemma}
    Instrument allocation can be modelled as in~\cite{RefWorks:RefID:1-čyras2019argumentation}, without optimality/efficiency AF.
    \begin{proof}
        From Definition \ref{def: instruments}, rule 1 can be modelled by the feasibility AF by its definition. Rule 2 can be modelled by mapping skill requirements to negative fixed decisions by Lemma \ref{lemma: violating skill requirements are negative fixed decisions}.
    \end{proof}
\end{lemma}
\begin{example}
\label{ex: instrument allocation}
    Consider two operators with skills:
    (1)  X Y Z,
        (2) Z, 
    and instruments $I0-I3$. Let $I1$ have skill requirements $\{X, Z\}$, and all other instruments be requirement-free.
    Let $$SI = \begin{pmatrix}
        1 & 1 & 0 & 0\\
        0 & 0 & 1 & 1
    \end{pmatrix}
    \quad SI' = \begin{pmatrix}
        1 & 0 & 0 & 0\\
        0 & 1 & 1 & 1
    \end{pmatrix}
    $$
    Then $SI$ is a feasible and valid instrument assignment, but $SI'$ is not since it violates the skill constraints for $I1$.
\end{example}

\begin{definition}
\label{def: job instrument assignment}
For $n$ jobs and $t$ instruments, we define $\zeta \in \{0,1\}^{n \times t}$ as the job-instrument constraint matrix, such that $\zeta_{j, \tau}$ is the element in the $j$th column and $\tau$th row of $\zeta$. We define $\zeta_{j, \tau}$ = 1 if instrument $\tau$ is a requirement for job $j$, and $0$ otherwise.
\end{definition}
\begin{definition}
    \label{def: job-instrument constraint AF}
    For $n$ jobs and $t$ instruments, consider job allocation schedule $S$ and instrument allocation schedule $SI$. Let $S_i$, $SI_i$ be the sets of jobs and instruments assigned to operator $i\in \mathcal{O}$ respectively.
    Then, the \textbf{job-instrument assignment AF} $(\mathit{Args_{JI}}, \rightsquigarrow_{JI})$ has:
    \begin{itemize}
        \item $\mathit{Args_{JI}} = \{a_{j, \tau} : j\in \mathcal{O}, \tau \in \mathcal{I}\}$,
        \item 
        $\rightsquigarrow_{JI} = \{(a_{j,\tau}, a_{j,\tau}): \tau \in SI_i, j \notin S_i, i \in \mathcal{O}
        \}.$
    \end{itemize}
\end{definition}
\begin{example}
\label{ex: instrument allocation af}
    Continuing from Example \ref{ex: instrument allocation}, suppose we also have jobs 1-4 with the following schedule assignment $S$ and job-instrument constraint matrix $\zeta$:
    \begin{equation*}
    S = \begin{pmatrix}
        1 & 0 & 0 & 1\\
        0 & 1 & 1 & 0\\
    \end{pmatrix} \quad \zeta = \begin{pmatrix}
        0 & 1 & 1 & 0\\
        0 & 0 & 0 & 0\\
        0 & 0 & 0 & 1\\
        0 & 0 & 0 & 0
    \end{pmatrix}.\end{equation*}
    Using $SI$ from Example \ref{ex: instrument allocation}, Figure \ref{fig: instrument allocation af} shows the job-instrument assignment AF with $\zeta$ highlighted in blue where $\zeta_{j, \tau} = 1$. There is a violation since $a_{1, I2} \rightsquigarrow a_{1, I2}$; $I2$ is an instrument requirement for job 1, however they are allocated to different operators.
    \begin{figure}[h]
        \centering
\includegraphics[width=0.47\linewidth]{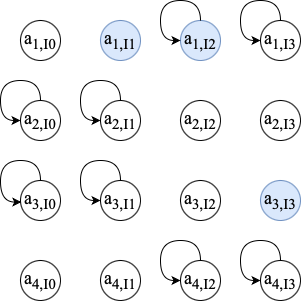}
        \caption{Job-instrument assignment AF for Example \ref{ex: instrument allocation af} }
        \label{fig: instrument allocation af}
    \end{figure}
\end{example}
\begin{lemma}
\label{lemma: time complexity job-instrument constraint AF}
    Given job and instrument assignment schedules $S, SI$ and job-instrument constraint $\zeta$, the job-instrument constraint AF can be constructed in $O(mnt)$ time, for $t$ the number of instruments.
\end{lemma}
\JL{The proof follows the same approach as Lemmas \ref{lemma: time complexity extended cost efficiency AF} and \ref{lemma: time complexity individual cost efficiency AF}}.
\section{Explanations}
\label{sec:explain}
We bring together the different argumentation frameworks to give comprehensive explanations for the validity of schedules. We extend \cite{RefWorks:RefID:1-čyras2019argumentation} to include two efficiency AFs (extended cost efficiency and individual schedule efficiency) and specify a subset of fixed decisions (skill requirements). In addition, we define an AF relating to fixed decisions for job-instrument allocation.
\begin{definition}
\label{def: extended explanations}
    For a given schedule $S$ and set of skills $\mathcal{K}$, let $E\approx S$ and $(\mathit{Args}, \rightsquigarrow)$ be any of $(\mathit{Args_F}, \rightsquigarrow_F), (\mathit{Args_{S+}}, \rightsquigarrow_{S+}), (\mathit{Args_{D_S}}, \rightsquigarrow_{D_S}), (\mathit{Args_{IS}}, \rightsquigarrow_{IS})\}$.
    For an attack $a \rightsquigarrow b$ with $ a, b \in E$:
    \begin{itemize}
        \item $(a, b) \in \rightsquigarrow_F \implies S$ is not feasible,
        \item $(a, b) \in \rightsquigarrow_{S+} \backslash \rightsquigarrow_F \implies S$ is not extended cost efficient,
        \item $(a, b) \in \rightsquigarrow_{D_S} \backslash \rightsquigarrow_F \implies S$ violates skill requirements,
        \item $(a, b) \in \rightsquigarrow_{IS} \backslash \rightsquigarrow_F \implies S$ is not individual cost efficient.
    \end{itemize}
    For a non-attack $E \cancel{\rightsquigarrow} b, b \notin E$:
        \begin{itemize}
        \item $\rightsquigarrow = \rightsquigarrow_F \implies S$ not feasible,
        \item $\rightsquigarrow = \rightsquigarrow_{S+}$ and $b \rightsquigarrow_{S+} E \implies S$ not efficient.
    \end{itemize}
Non-attacks are not possible for $(\mathit{Args_{D_S}}, \rightsquigarrow_{D_S}), (\mathit{Args_{IS}}, \rightsquigarrow_{IS})$ by definition.
\end{definition}
\begin{definition}
\label{def: job instrument explanations}
    For a given job-instrument constraint matrix $\zeta$, $E\approx \zeta$, $(\mathit{Args}, \rightsquigarrow) \in (\mathit{Args_{JI}}, \rightsquigarrow_{JI})$, for an attack $a \rightsquigarrow a, a \in E$:
        $(a, a) \in \rightsquigarrow_{JI} \implies$ job-instrument requirements are violated.
\end{definition}
\begin{theorem}
\label{th: total time complexity}
    Given $\mathcal{O}, \mathcal{J}, \mathcal{K}, \mathcal{I}$, schedules $S$, $SI$ and job-instrument constraint $\zeta$, the total time taken to construct the AFs for the problem is $O(mn(mn+K+t))$.
    Verifying explanations for the relevant extensions can be done in $O(m(mn^2 + t))$.
\end{theorem}
$SI$ and instrument-skills fixed decisions are derived from \cite{RefWorks:RefID:1-čyras2019argumentation}.
The proof of Theorem \ref{th: total time complexity} can be found in the Appendix.

\section{Evaluation}
\label{sec:eval}
{After optimal schedules are computed,} they may need to be modified over time to respond to last-minute changes. For example, workers could call in sick one day, or tasks may take longer than anticipated, causing a backlog. In these cases, coordinators must make manual changes in real time, however they may not be aware of how this impacts other areas of the schedule. They need to be able to explain how proposed modifications will lead to a new optimal schedule to ensure trust in the schedule.
We conducted a user study to compare how well the user is able to update an initial schedule, with and without our purpose-built tool, {utilising our explanation approach}.

Specifically, a user is given a schedule that needs modification as it is suboptimal and/or infeasible; they are tasked to produce an updated schedule. In this process, we evaluate the user’s speed and accuracy (the amount of time they need for producing the new schedule and the quality of the schedule produced).
The main finding is that our tool  supports \textit{Terranova} employees in obtaining modified schedules faster and of better quality{, with respect to not using our tool}.

\paragraph{Outline}
The user study 
used a proof-of-concept tool built upon \cite{AESwebapp}, implementing methodology from Sections~\ref{sec:theory} and \ref{sec:explain}.
Participants completed half of questions by hand and half aided by the tool
. Participants were randomly assigned one of four question sets, removing bias from the relative difficulty of questions and offering insight into how the question format impacted performance.

We compare performance across three key metrics: \textbf{schedule accuracy} (marked as correct if a user submitted a revised schedule which passed all feasibility and optimality checks from the tool, and incorrect otherwise), \textbf{cost accuracy} (marked as correct if a user's new calculated maximal cost across all operators matched that determined by the tool), and \textbf{completion time} (time taken for a user to complete a task, from the time the question was opened until the time an answer was submitted).

\paragraph{Method}
The study was conducted with 28 employees of 
\textit{Terranova}
. 
The user study consisted of four groups, who were each given two sets of questions with four distinct question orders:
\begin{itemize}
    \item Red (set 1 by hand; tool introduced; set 2 using tool),
    \item Yellow (set 2 by hand; tool introduced; set 1 using tool),
    \item Green (alternating questions from both sets with tool introduced at start; set 1 by hand, set 2 using tool),
    \item Blue (alternating questions from both sets with tool introduced at start; set 2 by hand, set 1 using tool).
\end{itemize}

\begin{figure}
    \centering
    \includegraphics[width=0.8\linewidth]{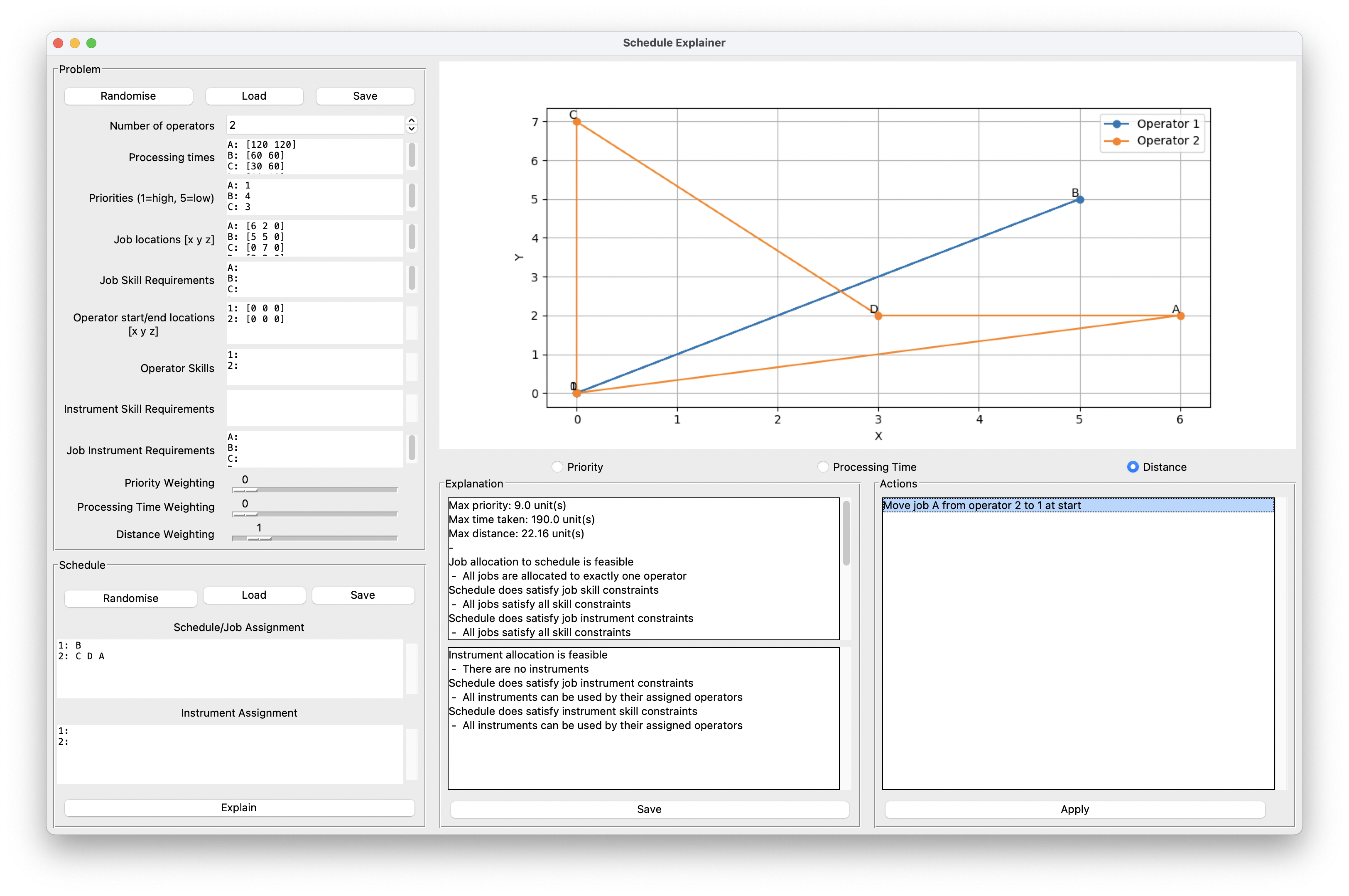}
    \caption{Tool interface with sample question from study}
    \label{fig: Sample question from user study}
\end{figure}

Questions ranged from optimising makespan or distance to optimising both, alongside instruments and skills constraints. For example, respondents were given the coordinates of jobs A(6,2), B(5,5), C(0,7), D(3,2). They were shown the schedule in Figure \ref{fig: Sample question from user study}, where operator 1 completes job B, and operator 2 completes the remaining jobs in the order C, D, A. Participants were then asked to optimise the schedules and give the correct cost, using the tool or by hand. A correct solution here would be to move C to operator 1.

The tool offered participants maximum flexibility to amend the inputs (through text boxes and sliders) and select different optimisation suggestions from the tool's output. Diagrams charting travel distance and processing time  allowed participants to visualise the impact of schedule changes, before selecting from the list the tool provided.
\begin{figure*}
\includegraphics[width=\linewidth]{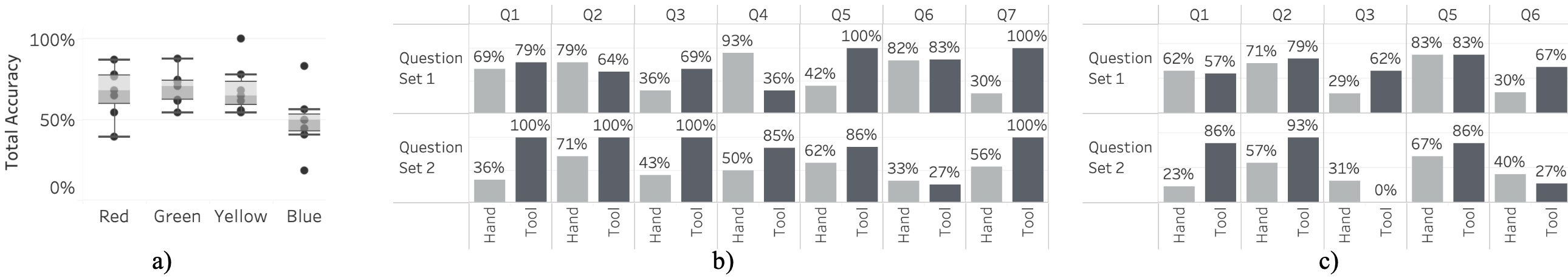}
  \caption{Accuracy results. a) Overall accuracy by survey group; b) Schedule accuracy for question sets 1 and 2; c) Accuracy of maximum cost responses.}
  \label{fig: user study all accuracy charts}
\end{figure*}

There was significant positive correlation between engagement rate (number of questions answered, compared to those left blank) and total accuracy
(all schedule and cost questions answered correctly), $r(26) = -.37, p=.026$.  Figure \ref{fig: user study all accuracy charts}a 
shows that accuracy was mostly consistent across survey groups; retention rate dropped for the last, with 
{only} half of respondents reaching the end.
\paragraph{Schedule accuracy}
Respondents were asked to provide an improved schedule for each problem (or for question 4, respondents picked the optimal schedule from options given). For questions 1-3 and 5-6, respondents included the maximum distance or makespan for their optimised schedule. This was omitted from question 4, for obvious reasons, and question 7 as it utilised a combination of distance and time (making the question ambiguous).

Accuracy is calculated as the {number of correct responses out of all users that give a response}
. The following was assumed: (1) schedules were counted as correct if the tool identifies them as feasible, efficient, individually efficient and all fixed decisions are met; (2) the maximum cost for any operator was correct (regardless of whether the tool said the schedule given is efficient or not); (3) for questions involving instrument allocation, we assumed (where not given) the respondent found a correct instrument allocation, if one existed.

Figure~\ref{fig: user study all accuracy charts}b shows the proportion of correct answers for each schedule question, split into respondents who answered by hand versus those who used the tool.
We observed that, in general, respondents who used the tool were more likely to get a question correct, except for
    (1) Set 1, Q4 (Choice of Schedule): it is unclear why respondents struggled to use the tool to answer this question – possibly, having multiple schedules to download and compare may have caused some confusion, as the answer was relatively easy to deduce by analysing the three schedules (route optimisation/distance) visually; 
        (2) Set 2, Q6 (Distance with Instruments): this problem introduced instrument assignments and constraints
        leading to the tool creating a loop (job $F$ required instruments I1 and I0, but these were allocated to different operators, so moving the job did not help in solving the problem) - the correct answer here was to move one of two instruments, so that both instrument requirements for job $F$ were met, then optimise.       
        This was the worst answered question out of all those in the user study, with just 29\% of respondents giving a correct schedule.
\paragraph{Cost accuracy}
Figure~\ref{fig: user study all accuracy charts}c shows the proportion of correct answers for each question with numerical input (max cost, for distance and processing time problems), split into respondents who answered by hand versus by tool. Note that, for these questions, we only accepted the true optimal value (i.e. a lowest max cost given, where the schedule answer provided passed all tests by the tool and gave the same cost value).
We observed that 
results were generally much more variable than the schedule questions: this was likely due to the tool's sub-optimal algorithm, meaning some solutions found by hand were better than solutions immediately found by the tool.

There were some anomalies.
(1) Set 1, Q1 (Distance): despite the tool giving the KPIs, a significant proportion of respondents using the tool put ``13" for the distance instead of ``26" for the distance travelled (not counting the return journey); regardless, these respondents gave a correct schedule solution.
(2) Set 2, Q3 (Time): as discussed above, this was a question where the tool recommended a schedule that was sub-optimal. It is worth noting that every respondent here gave a makespan answer which was optimal by the tool's standards.
(3) Set 2, Q6 (Distance with Instruments): as with the schedule response, this question had a low success rate with the tool due to lack of understanding applying changes to the instrument allocation.

Still, for the remaining questions, the accuracy rate using the tool was greater than that completing the questions by hand.
{Across all questions, the accuracy rate was 54\% by hand, compared to 74\% for questions answered with aid of the tool. There was significant improvement in accuracy when respondents were able to use the tool, $\chi^2 (1, N=608) = 24.95, p<10^{-4}$.}
\paragraph{Completion Time}
The average completion time for the user study was 2 hours and 23 minutes. Despite minor variations in time completion depending on which question set was completed by hand versus the tool (2 hours 29 minutes vs 2 hours 19 minutes), respondents who alternated generally took longer (2 hours 51 minutes compared to 1 hour 57 minutes). The difference may be due to respondents not becoming sufficiently used to completing questions with the tool or by hand, due to constant switching.
\begin{figure}
  \centering
\includegraphics[height=0.35\linewidth]{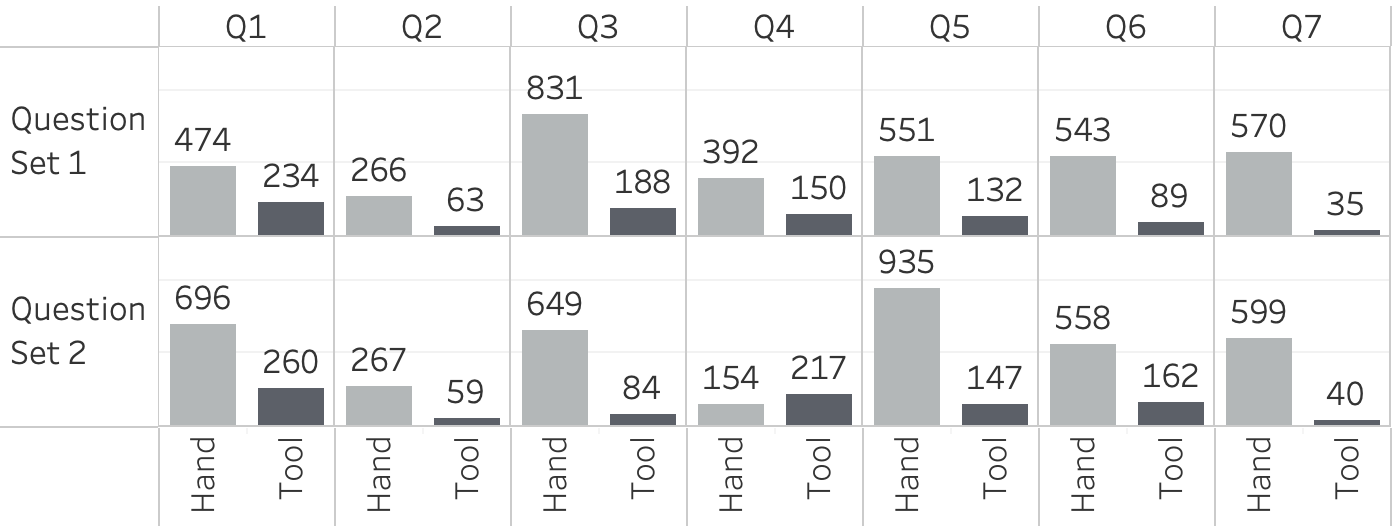}
 \caption{Average time completion (seconds) per question}
 \label{fig: Average time completion for question sets}
\end{figure}
Figure \ref{fig: Average time completion for question sets} 
    shows the average time spent on each question. We omit values where the user did not submit a response (as we cannot tell whether users attempted the question before skipping it).
We
observe:
(1)  the first question in both problem sets had a high average completion time, likely due to lack of familiarity with the format and/or tool;
(2) Q2 from both sets were simple makespan problems, while Q3 used much larger examples (around double the number of jobs); this was reflected in higher completion times, particularly for those completing the problems by hand;
(3) remaining questions were intended as medium difficulty (albeit challenging to solve by hand), with the exception of Q7, which was solved quickly by respondents using our tool, likely due to users now being familiar with the tool and survey format; unlike previous questions, this question contained no constraints (skills or instruments).

{On average, the median time taken for users to complete questions by hand was more than 4 times slower than with aid of the tool (368 seconds vs 71 seconds). There is very strong positive correlation indicating that a question could be solved faster using the tool $(M=135.8, SD=187.5)$, compared to by hand $(M=527.5, SD=557.2), t(354)=9.04, p<10^{-17}$.}
\section{Conclusions}
We have introduced a methodology for equipping workforce management optimisation with explanations and tested the methodology with a user study showing clear improvements in accuracy and efficiency in an industry setting. This shows promise in effectiveness with identifying and revising solutions to the problem.  

As future work, we plan to accommodate argumentation for sequencing of jobs as well as to conduct  further, more extensive, user studies towards full deployment of our methodology in industry settings. It would also be interesting to  explore 
other ways to present explanations to users as well as the use of argumentation for other optimisation settings.



\bibliography{p879}


\appendix

\begin{center}
{\bf\large Argumentation for Explainable Workforce Optimisation:
Appendix}  
\end{center}

\label{app:proofs}




\paragraph{Proof of Lemma~\ref{lemma: Every optimal schedule with respect to extended cost satisfies SEP+ and PEP+}}
    Let $S$ be the original optimal schedule.
        SEP+ (the proof for PEP+ is similar): 
        \JL{Consider another operator $i’$ and two consecutively executed jobs $j'$ and $j'^+$.}
        Let $S*$ be the schedule obtained by moving job $j$ from machine $i$ to $i'$, placed directly after $j'$, keeping the remaining job assignments and orders as in $S$ (i.e. $x_{i',j,k'+1}=1$). \JL{If we put $j$ between $j'$ and $j'^+$, then we shouldn’t get a better schedule.}

Then the new costs of $i, i'$ will be $C_i(S^*) = C_i(S) - \alpha p_{i,j} - \beta [d_{j^-, j} + d_{j, j^+}] + \beta d_{j^-, j^+}$ and $C_{i'}(S^*) = C_{i'}(S) + \alpha  p_{i',j} + \beta [d_{j', j} + d_{j, j'^+}] - \beta d_{j', j'^+}$. Note $C_{i''}(S^*) = C_{i''}(S)$ for all $i'' \neq i, i'$.

Since job $j$ is critical, $i$ is the critical operator and so $C_{i}(S) \geq C_{i'}(S)$. Since $S$ is optimal, we have that $S^*$ cannot obtain a lower extended cost and so $C_{i'}(S^*) \geq C_i (S)$. Putting this together: $C_{i'}(S^*) \geq C_i (S)$

    $\implies C_{i'}(S) + \alpha p_{i',j} + \beta [d_{j', j} + d_{j, j'^+}] - \beta d_{j', j'^+} \geq C_i (S)$
    
    $\implies C_i (S) - C_{i'}(S) \leq \alpha p_{i',j} + \beta [d_{j', j} + d_{j, j'^+}] - \beta d_{j', j'^+}.$

  \paragraph{Proof of Theorem~\ref{theo:feas}}
    
        This proof is derived from Theorem 4.3 from \cite{RefWorks:RefID:1-čyras2019argumentation}.
        Let $E$ be stable in $(\mathit{Args_{S+}}, \rightsquigarrow_{S+})$.
        We know $E$ is conflict-free in $(\mathit{Args_{F}}, \rightsquigarrow_{F})$ since attacks removed for SEP+ only make asymmetric attacks symmetric and attacks added for PEP+ are between arguments that do not already attack each other in $(\mathit{Args_{F}}, \rightsquigarrow_{F})$ (since $i \neq i'$).
        So $S$ is feasible by Theorem 4.1 in \cite{RefWorks:RefID:1-čyras2019argumentation}.

    As $E$ is stable for all $j \in \mathcal{J}, a_{i,j} \in E$, it follows that $ a_{i,j} \rightsquigarrow_{S+} a_{i',j}$ where $i' \neq i$. So we do not need to remove attacks from $\rightsquigarrow_{F}$ to get to $\rightsquigarrow_{S+}$, so $\forall i\neq i', j \neq j', x_{i,j,k} = x_{i',j',k'}=1: 
        C_i - C_{i'} >\alpha p_{i',j} + \beta [d_{j', j} + d_{j, j^+}] - \beta d_{j', j^+}$ cannot hold for any $(i, i', j)$, leading to SEP+ being satisfied.

For PEP+, since $E$ is conflict-free, $\forall i\neq i', j \neq j', x_{i,j,k} = x_{i',j',k'}=1,$ 
        
        $\beta ([d_{j^-, j} + d_{j, j^+}] - [d_{j^-, j'} + d_{j', j^+}]) > \alpha [p_{i, j'}-p_{i,j}], $
        
         $ C_i - C_{i'} >
\alpha [p_{i', j}-p_{i',j'}] + \beta ([d_{j'^-, j} + d_{j, j'^+}] - [d_{j'^-, j'} + d_{j', j'^+}])$
\\
cannot hold for any $(i, i', j, j')$ so $S$ satisfies PEP+.

As in Theorem 4.3 from \cite{RefWorks:RefID:1-čyras2019argumentation}, since $S$ is feasible, $E$ is stable in $(\mathit{Args_{F}}, \rightsquigarrow_{F})$. $S$ also satisfies SEP+ and PEP+ so $\rightsquigarrow_{S+} = \rightsquigarrow_{F}$. So $S$ feasible, satisfies SEP+, PEP+ $\implies$ $E$ stable in $(\mathit{Args_{S+}}, \rightsquigarrow_{S+})$.

   \paragraph{Proof of Lemma~\ref{lemma: time complexity extended cost efficiency AF}}
        Recall that there may be multiple $i \in \mathcal{O}$ that satisfy $C_{max} = C_i$. Recall from 
        Lemma 4.2
        in \cite{RefWorks:RefID:1-čyras2019argumentation} the construction of the feasibility AF takes $O(nm^2)$. To remove attacks, we must first identify the critical operators, taking $O(m)$. Then, for each critical operator (of which there are a maximum of $m$), to iterate over $i \in \mathcal{O}, j,j' \in \mathcal{J}$ and check $j\neq j'$, $x_{i,j,k} = x_{i', j', k'} = 1$ takes $O(mn^2)$. For each of these, calculating the distance metrics and the inequality comparisons take $O(1)$. So, in total, removing attacks takes $O(m^2n^2)$.
        To add attacks, we iterate over $i \in \mathcal{O}, j,j' \in \mathcal{J}$ for each critical operator and check $x_{i,j,k} = x_{i',j',k'}=1, i \neq i', j \neq j'$ which takes $O(m^2n^2)$.
        Calculating distances ($O(1)$) and checking inequalities ($O(1)$) means the addition of attacks  takes $O(m^2n^2)$.
    As with Lemma 4.2
    in \cite{RefWorks:RefID:1-čyras2019argumentation}
    , we determine if $E \subseteq Args_{S+}$ is stable by checking if $E$ is conflict-free, taking $O(m^2n^2)$, and if $E$ attacks every argument in $Args_{S+} \backslash E$, also taking $O(m^2n^2)$.

\paragraph{Proof of Lemma~\ref{lemma:optimal schedule satisfies ISEP IPEP}}
    ISEP: Let $D_i(S)$ be the objective distance function for a schedule $S$, where
    $$D_i = \sum_{j,j'}y_{i,j,j'}d_{j,j'} + \sum_{j} x_{i,j,0} d_{0,j} + \sum_{j} z_{j} d_{j,0}.$$
    Let S* be an optimal schedule. If we move job $j$ from its position between $j^-$ and $j^+$ to a new position between $w$ and $w^+$, we have
    \begin{align*}
        D_i(S) = D_i(S*) - [d_{j^-, j} + d_{j, j^+}] +  d_{j^-, j^+} \\
        + [d_{j', j} + d_{j, j'^+}] - d_{j', j'^+}.
    \end{align*}
    Since $S*$ is optimal, any other schedule route will be longer and such $D_i(S) \geq D_i(S*)$. 
    The proof for IPEP is similar.

\paragraph{Proof of Theorem~\ref{th: total time complexity}}

        For $S$, Lemmas 4.2  in \cite{RefWorks:RefID:1-čyras2019argumentation}, \ref{lemma: time complexity extended cost efficiency AF}, \ref{lemma: time complexity skill constraint AF}, \ref{lemma: time complexity individual cost efficiency AF} and \ref{lemma: time complexity job-instrument constraint AF} give all constructions/explanations relating to job assignment. For $SI$, Lemmas 4.2 in \cite{RefWorks:RefID:1-čyras2019argumentation} and \ref{lemma: time complexity skill constraint AF} give constructions relating to instrument assignment. In Lemma \ref{lemma: time complexity job-instrument constraint AF}, checking for self-attacks can be done in $O(nt)$ time. 
        Checking stability for the remaining AFs, the theorem follows.
   
\end{document}